\documentclass{article}

     \PassOptionsToPackage{numbers, compress}{natbib}

\usepackage[preprint]{neurips_2024}

\usepackage[utf8]{inputenc} 
\usepackage[T1]{fontenc}    
\usepackage{hyperref}       
\usepackage{url}            
\usepackage{booktabs}       
\usepackage{amsfonts}       
\usepackage{nicefrac}       
\usepackage{microtype}      
\usepackage{xcolor}         
\usepackage{amsmath}
\usepackage{enumerate}
\usepackage{graphicx}

\title{Network Representation Learning for \\ Biophysical Neural Network Analysis}

\author{%
  Youngmok Ha\footnotemark[1] \\
  ETRI\\
  Daejeon, Republic of Korea \\
  \texttt{ymha@etri.re.kr} \\
  \And
  Yongjoo Kim \\
  ETRI\\
  Daejeon, Republic of Korea \\
  \texttt{y.kim@etri.re.kr} \\
  \And
  Hyun Jae Jang \\
  KIST \\
  Seoul, Republic of Korea \\
  \texttt{hjjang}@kist.re.kr \\
  \And
  Seungyeon Lee \\
  KIST \& Korea University\\
  Seoul, Republic of Korea \\
  \texttt{seungyounlee}@kist.re.kr \\
  \And
  Eunji Pak \\
  ETRI \\
  Daejeon, Republic of Korea \\
  pakeunji@etri.re.kr \\
}

\begin{document}

\maketitle

\renewcommand{\thefootnote}{\fnsymbol{footnote}}
\footnotetext[1]{Corresponding author.}

\begin{abstract}
The analysis of biophysical neural networks (BNNs) has been a longstanding focus in computational neuroscience. A central yet unresolved challenge in BNN analysis lies in deciphering the correlations between neuronal and synaptic dynamics, their connectivity patterns, and learning process. To address this, we introduce a novel BNN analysis framework grounded in network representation learning (NRL), which leverages attention scores to uncover intricate correlations between network components and their features. Our framework integrates a new computational graph (CG)-based BNN representation, a bio-inspired graph attention network (BGAN) that enables multiscale correlation analysis across BNN representations, and an extensive BNN dataset. The CG-based representation captures key computational features, information flow, and structural relationships underlying neuronal and synaptic dynamics, while BGAN reflects the compositional structure of neurons, including dendrites, somas, and axons, as well as bidirectional information flows between BNN components. The dataset comprises publicly available models from ModelDB, reconstructed using the Python and standardized in NeuroML format, and is augmented with data derived from canonical neuron and synapse models. To our knowledge, this study is the first to apply an NRL-based approach to the full spectrum of BNNs and their analysis.
\end{abstract}

\section{Introduction}

The analysis of biophysical neural networks (BNNs) is a central focus of computational neuroscience, as it advances our understanding of neurological functions and their applications. By examining the structural and dynamic properties of neural networks, BNN analysis reveals how these complex systems influence a wide range of functions, from sensory processing to higher cognitive activities. These insights are essential for understanding neural processing, interconnectivity, and learning. They also enable the reconstruction and simulation of neural architectures, thereby driving advancements in brain-like computing, such as neuromorphic systems~\citep{roy2019towards} and bio-inspired intelligence models~\citep{rumelhart1986learning,lecun1998gradient,maass2002real}.

BNN analysis presents challenges due to the high complexity of the networks. The complexity stems from the diverse anatomical, biochemical, electrophysiological types of neurons, including sensory, motor, and interneurons. The intricate connectivity patterns established by these neurons, coupled with dynamic synaptic plasticity mechanisms, further amplify their complexity. Moreover, the dynamic nature of BNNs, characterized by continuous activity-dependent changes and adaptive behaviors, adds additional layers of difficulty in deciphering their organizational structures and functional properties.

Extensive studies has been conducted to unravel the complexity, with efforts grounded in biophysical and neurobiological principles. Foundational work by pioneers such as Lapicque~\citep{louis1907lif} and Hodgkin \& Huxley~\citep{hodgkin1952quantitative} has been expanded by subsequent studies~\citep{hasselmo1994laminar,markram1996redistribution,tsodyks1997neural,izhikevich2003simple,lisman1999relating,gerstner2002spiking,koch2004biophysics,leutgeb2005independent,rolls2006computational,izhikevich2007dynamical,mongillo2008synaptic,aimone2009computational,gupta2010hippocampal,giraud2012cortical}. These collective efforts have advanced the study of more refined BNNs and their learning processes~\citep{eliasmith2012large,yang2019task,gallego2020event,yang2020artificial,fang2023parallel}. 
Simultaneously, computational approaches leveraging simulation and virtualization have enhanced the capacity to analyze the complexity of BNNs ~\citep{hines1997neuron,diesmann2001nest,stimberg2019brian,gleeson2010neuroml,sanz2013virtual,furber2014spinnaker,niedermeier2022carlsim,akopyan2015truenorth,davies2018loihi,dura2019netpyne,pehle2022brainscales,richter2024dynap}.  

Despite these advances, several crucial questions in BNN analysis remain unresolved. Among them, a fundamental challenge is understanding the precise correlations between the dynamic properties of neuronal or synaptic components in BNNs, their connectivity patterns, and the associated learning processes. For instance, our understanding of how specific dynamic properties, such as excitability, directly influence learning activities is still incomplete. Moreover, the relationship between structural changes in neural networks during learning and established learning rules remains unclear. Additionally, the nonlinear and heterogeneous interactions between BNN components across multiple scales, from local neuronal and synaptic dynamics to global network behaviors, require deeper investigation.

One main reason for these unresolved issues is the absence of a comprehensive and multifaceted approach. Unraveling the correlations requires a methodology that integrates dynamic modeling, structural tracking, and multiscale interaction analysis, all while ensuring interpretability.

To tackle this demand, we propose a novel BNN analysis framework based on network representation learning (NRL)~\citep{zhang2018network}, which provides insights into correlations between network components via attention scores. The contributions of our NRL framework can be summarized in three key aspects. First, it introduces a new BNN representation based on computational graphs (CGs), which captures computational features, information flow, and structural relationships related to neuronal and synaptic dynamics. Second, it proposes a novel bio-inspired graph attention network (BGAN) specifically designed to enable multiscale understanding of correlations within BNN representations. Lastly, it addresses the lack of data required for the NRL of BNNs.


\section{Related Work}

To the best of our knowledge, this work is the first to pioneer an NRL-based study of a broad range of BNNs. Previous studies, such as \citep{li2023uncovering,zhang2024enhancing}, have applied NRL to spiking neural networks based on the leaky integrate-and-fire model, which represents only a subset of the broader BNN family. Other works~\citep{parisot2017spectral,ktena2018metric,parisot2018disease,li2019graph,jiang2020hi,tong2023fmri} have focused on analyzing structural or functional connectivity among brain regions. However, these works target higher-level brain networks, rather than addressing BNNs with detailed biophysical mechanisms at the level of individual neurons and synapses. Consequently, no prior work has addressed NRL for the full BNN family, resulting in a lack of the three essential NRL components: robust BNN representations, specialized learning models, and sufficiently detailed data. This work addresses these gaps by introducing a novel NRL framework that integrates CG-based BNN representation, BGAN, and BNN dataset.


\section{Background Knowledge}
This paper requires knowledge of BNNs, CGs, and self-attention mechanisms. We provide the basic concepts in Appendix~\ref{sec_bk} due to the page limitation.

\section{Overview}
\textbf{Goal.} This work aims to unravel the correlations between the dynamic properties of BNN components, their connectivity patterns, and the associated learning processes by pioneering the NRL of BNNs.

\textbf{Methodology.} We leverage NRL~\citep{zhang2018network}, a widely adopted approach across various domains for providing numerical insights into correlations between network component properties, capturing structural relationships within networks, and enabling scalable processing~\citep{wang2016structural,tu2018structural,tu2018deep,gao2018deep,velivckovic2018graph,zheng2019multimodal,zhou2020transferable,gagrani2022neural}. For the NRL of BNNs, we propose a new BNN representation based on CGs and BGAN that incorporates novel neural structural attention (NSA) and bidirectional masked self-attention (BMSA) mechanisms. In the absence of a standardized dataset, we construct a new BNN dataset to facilitate learning.

\textbf{Rationale.} The rationale for adopting CGs to represent BNNs lies in the dynamic nature of these networks, where neurons and synapses function as computational units with diverse features related to neuronal or synaptic dynamics. Simple graphs used in previous studies are insufficient to capture the full range of functionality and diversity in these computational units. In contrast, our CG-based BNN representation effectively models these features, alongside the dynamic flow of information and structural relationships within the network.

BGAN, along with its NSA and BMSA mechanisms, is inspired by the fundamental characteristics of BNNs. NSA reflects the hierarchical structure of neurons, composed of compartments such as dendrites, soma, and axons. BMSA is motivated by the bidirectional nature of signal transmission in BNNs, where communication occurs in both forward and backward directions. NSA tightly couples neurons and their compartments in a deterministic manner. Since both NSA and BMSA are derived from attention mechanisms~\citep{bahdanau2015neural,vaswani2017attention}, their attention scores help to uncover correlations within our CG-based BNN representation from both local and global perspectives.

The availability of ModelDB~\citep{hines2004modeldb}, which serves as a central hub for researchers in neuroscience and brain science to share and reproduce neural models globally, provides the foundation for gathering our dataset. From ModelDB, we collect publicly available BNN models. The models in ModelDB exhibit diversity in their focus, representation methods, and simulation environments, depending on the research objectives. Therefore, we standardize the collected models. For standardization, we use NeuroML~\citep{gleeson2010neuroml}, which provides a common framework for defining network morphology and connectivity, ensuring consistency in the representation and comparison of diverse models. To augment the limited dataset, we also incorporate synthetic data.


\section{Framework}
\begin{figure}[ht]
    \centering
\includegraphics[width=0.98\linewidth]{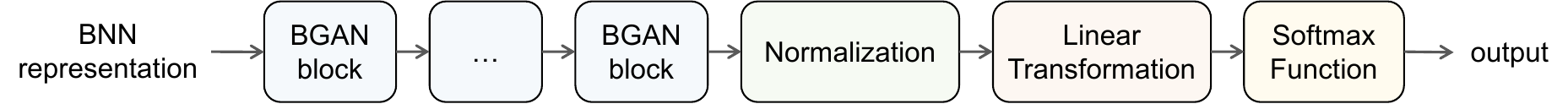}
    \caption{Processing flow of BNN representation through the BGAN block in the proposed framework}
    \label{fig:fr}
\end{figure}

We introduce a novel NRL framework for BNN analysis, grounded in the CG-based BNN representation, BGAN, and BNN dataset. Within our NRL framework, the CG-based BNN representation is processed through the BGAN block as illustrated in Figure~\ref{fig:fr}. When multiple BGAN blocks are employed, the output from each block serves as the input for the next, progressively refining the representation. Each BGAN block has distinct learnable parameters, allowing it to capture different aspects of the representation. The final output from the BGAN block is processed through normalization, linear transformation, and softmax operations, similar to those in language models~\citep{vaswani2017attention,devlin-etal-2019-bert,brown2020language,touvron2023llama}.

\subsection{BNN Representation}
We define a new BNN representation by treating BNN components such as neurons, dendrites, somas, axons, and synapses as computational nodes within computational graphs (CGs). This approach is justified by the fact that each of these components performs computations related to neuronal or synaptic dynamics, as described in Appendix~\ref{bnn}. 

The core task in adopting CGs is to design the features of both nodes and edges. Our computational nodes are characterized by six common features:
\begin{enumerate}[(i)] 
    \item \textbf{Node type}, such as neuron, dendrite, soma, axon, or synapse, which is used for structuring the CG representation and clarifying biophysical semantics,
    \item \textbf{Node index}, a non-negative integer value to facilitate the fast location of specific nodes in CG,
    \item \textbf{List of input/output node properties}, including node type and index, to include information about the connections between nodes,
    \item \textbf{List of geometrical information}, such as length, width, and positional information, to enable for the simple handling of spatial information, thereby reducing the computational cost of the NRL compared to NRL in three-dimensional spatial space, 
    \item \textbf{List of activity information}, such as input/output times of electrical or chemical signals, firing times, and firing patterns, which has information on signal communication and processing between nodes, and 
    \item \textbf{List of functional information}, including source code for dynamics implementation, as well as abstract syntax trees, data or control flow graphs of the source code, which contains detailed representations of neuronal or synaptic dynamics, including the mathematical models, algorithms, and parameters that define how nodes behave over time, interact with each other, and respond to inputs.
\end{enumerate}
Additionally, synapse nodes include the feature of synaptic efficacy, while neuron nodes have the feature of compartmental information. The edges between computational nodes are designed to include inter-node signal communication statistics.

It is noteworthy that our CG-based BNN representation manages a list of functional information. The source code specifies various components of neuronal or synaptic dynamics, such as membrane potentials, ion channel activity, synaptic transmission, and plasticity mechanisms. The abstract syntax tree (AST) provides a hierarchical representation of the source code’s syntax, aiding in structural analysis and optimization. The data flow graph (DFG) and control flow graph (CFG) offer more detailed insights into the structure and behavior of these dynamics. While the DFG focuses on how data is processed, the CFG emphasizes the logical flow of control within the dynamics. 

For example, the DFG illustrates how variables such as membrane potentials, ion channel currents, and gating variables progress through different stages of computation. It also highlights the dependencies between operations, offering insight into how data is processed and transformed within the code. Furthermore, the CFG represents the sequence of operations and the branching structure of the code, determining how different parts of the neuronal and synaptic models are executed based on conditions such as synaptic connections, inputs, membrane potential values, and ion channel properties.

\subsection{BGAN Architecture}

\subsubsection{Neuronal Structural Attention (NSA)}
The BGAN block incorporates NSA by hierarchically combining two graph attention network (GAN) blocks, $\text{GAN}_{low}$ and $\text{GAN}_{up}$, where $low$ and $up$ denote the lower and upper levels of the neuronal structural hierarchy, respectively. In BNNs, dendrites, somas, and axons comprise neurons, which are connected via synapses. Thus, $\text{GAN}_{low}$ manages nodes representing dendrites, somas, and axons, while $\text{GAN}_{up}$ handles nodes representing neurons and synapses. The hierarchical structure of NSA offers the advantage of coupling neurons and their compartments in a deterministic manner.

\begin{figure}[ht]
    \centering
\includegraphics[width=0.99\linewidth]{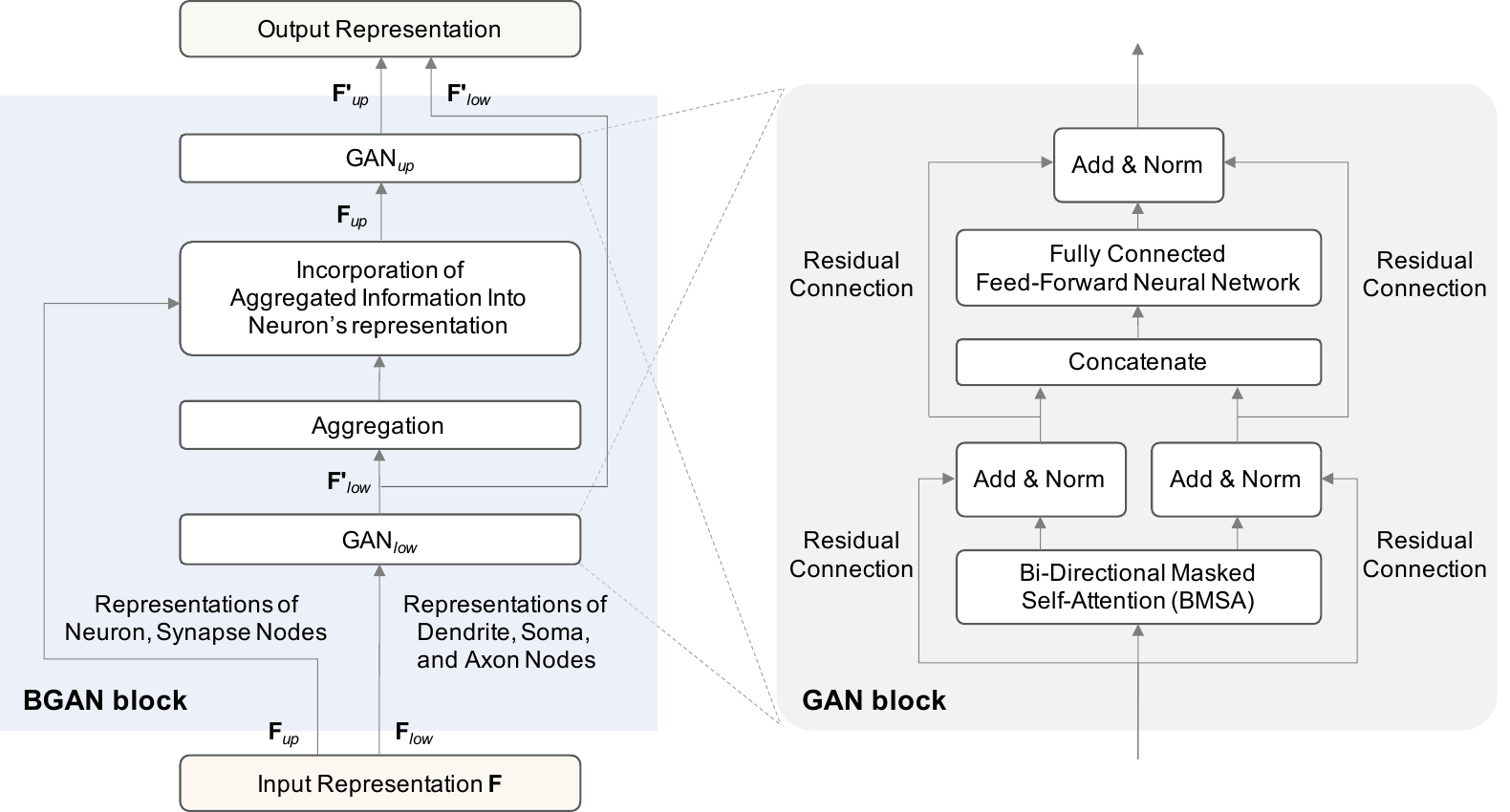}
    \caption{Architecture of the BGAN and GAN block: The BGAN and GAN block process input representations via NSA and BMSA, respectively.}
    \label{fig:model_arch}
\end{figure}

Figure~\ref{fig:model_arch} describes how NSA of the BGAN processes the input representation $\mathbf{F}=\{\mathbf{F}_{low},\mathbf{F}_{up}\}$, where $\mathbf{F}_{low}$ and $\mathbf{F}_{up}$ denote the input representations of \{dendrites, somas, axons\} and \{neurons and synapses\}, respectively. First, $\text{GAN}_{low}$ processes $\mathbf{F}_{low}$. The output of $\text{GAN}_{low}$ forms $\mathbf{F}'_{low}$, part of the BGAN block's overall output. Since the output of $\text{GAN}_{low}$ includes information about neuron's compartments, the information, or part of it, is aggregated and incorporated into the neuron's representation. Subsequently, $\text{GAN}_{up}$ processes $\mathbf{F}_{up}$, generating another component of the BGAN block's output $\mathbf{F}'_{up}$. 

$\text{GAN}_{low}$ and $\text{GAN}_{up}$ may have separate learnable parameters, and both consist of BMSA, residual connections, matrix summations, normalization, and a fully connected feed-forward network, with all components except BMSA functioning similarly to those in language models.

\subsubsection{Bidirectional Masked Self-Attention (BMSA)}
BMSA is motivated by the structural connections and information flow between BNN components. 

In BNNs, the typical flow of information signals follows the direction from the presynaptic neuron $\rightarrow$ synapse $\rightarrow$ postsynaptic neuron, and from the dendrites $\rightarrow$ soma $\rightarrow$ axon, where neurons are connected via synapses and consist of dendrites, soma, and axon. However, information signals can also flow in the opposite direction due to mechanisms such as backpropagating action potentials, retrograde signaling, dendritic action potentials, and synaptic plasticity. 

BMSA employs direction-specific masking strategies to manage connections and information flow in both directions, where masking refers to the method of blocking or ignoring certain positions in the input vectors or their elements during attention calculations. The direction-specific masking strategies utilize two types of masking functions, $\text{Mask}_A$ and $\text{Mask}_B$, where $A$ and $B$ denote forward and backward, respectively. Since $\text{Mask}_A$ and $\text{Mask}_B$ reflect structural connections, they incorporate a directed connection masking strategy~\cite{velivckovic2018graph}. Additionally, to account for the direction of information flow, they operate based on causal attention masking~\cite{vaswani2017attention} and reverse causal attention masking, respectively.

Based on $\text{Mask}_A$ and $\text{Mask}_B$, BMSA outputs two attention results, $\mathbf{F}'_A$ and $\mathbf{F}'_B$, for a given input representation matrix $\mathbf{F}$. Suppose we have already computed the \{Query,Key,Value\} matrices $\{\mathbf{Q}, \mathbf{K}, \mathbf{V}\}$ from $\mathbf{F}$, and a score matrix $\mathbf{C}$ (see Appendix~\ref{sam} for the computation). The output attention results are $\mathbf{F}'_A=\text{Softmax}(\text{Mask}_A(\mathbf{C})) \cdot \mathbf{V}$ and $\mathbf{F}'_B=\text{Softmax}(\text{Mask}_B(\mathbf{C})) \cdot \mathbf{V}$, respectively.

\subsubsection{Input Representation}

We describe the method for constructing the input representation $\mathbf{F}$, as it impacts computational efficiency, interpretability, model performance, and more. To construct $\mathbf{F}$, we base it on the proposed CG representation. Let $\mathbf{f}_{CG,i}$ denote the feature vector of the $i$-th computational node in the CG. The feature vector $\mathbf{f}_{i}$, a component of $\mathbf{F}$, is generated by prepending a special token [CLS], which encapsulates the summary information of each node, to $\mathbf{f}_{CG,i}$. This approach draws inspiration from the BERT methodology~\citep{devlin-etal-2019-bert}, which employs [CLS] to aggregate sentence pairs into a single sequence for processing in language tasks.

The method of constructing $\mathbf{F}$ from $\mathbf{f}_{i}$ depends on the attention type of NRL: node-wise or feature-wise. For node-wise attention, we construct $\mathbf{F}$ by stacking the transposed vectors, $\mathbf{f}_i^T$. For feature-wise attention, $\mathbf{F}$ is constructed by stacking $\mathbf{f}_i$ and interspersing another special token [SEP] to segment individual nodes. In this work, we employ node-wise attention to investigate the correlations between structural aspects and learning processes. Additionally, we apply feature-wise attention to study the correlations between properties of neuronal or synaptic dynamics and learning processes, as well as the interactions among BNN components.

\subsection{BNN Dataset}
\subsubsection{Real Data}
For decades, numerous biophysical models have been developed, covering various brain regions, different neuron types, synapse models, and synaptic plasticity mechanisms. The models aim to replicate the brain's complex dynamics by simulating its behavior at various levels of biological detail. As of September 7, 2024, statistics from ModelDB~\citep{hines2004modeldb} (\href{URL}{https://modeldb.science/trends}) indicate that approximately 1,870 publicly available models have been developed across a range of simulation environments. 

From ModelDB, we select models that focus on forming network structures, rather than individual neurons or synapses. We then reconstruct and standardize the selected models due to the variability in their specific focus, model representation methods, and simulation environments. In each model, synapse-related objects are identified to determine the presynaptic and postsynaptic neurons. Subsequently, we reconstruct each neuron by modeling its compartmental sections, with each section incorporating the necessary ionic mechanisms to compute membrane potentials. For standardization, we use NeuroML~\citep{gleeson2010neuroml}, a widely adopted format for encoding neural models. 

Furthermore, we evaluate the responses of the reconstructed networks by comparing them with those of the original networks under a common task. Poisson-distributed spike patterns, ranging from 0 to 50 Hz in 5 Hz increments, is applied, and the mean firing rate of each neuron is calculated. Simulations of the standardized network models is performed using the NetPyNE library~\citep{dura2019netpyne}, built on the NEURON simulator~\citep{hines1997neuron}, operating at a simulation frequency of 10 kHz. This approach ensures a controlled environment, enabling consistent comparisons of network behavior across different BNN models.

\subsubsection{Synthetic Data}
We also generate synthetic data to augment the dataset by employing the standard neuron and synapse models described in Section \ref{bnn}. This involves randomly sampling parameters such as the number of compartments, neurons, synapses, and their dynamic properties, and generating source code to implement them. To further enrich the synthetic data, we train synthetic complex BNN models and simple SNN models using spike-timing-dependent plasticity~\citep{feldman2012spike} or surrogate gradient descent~\citep{neftci_etal19}.


\section{Ongoing and Future Works}
We are currently enhancing the proposed NRL framework through pre-training tasks, which include autoencoding, masked representation modeling and edge prediction following methods described in~\citep{guo2021graphcodebert}. We track and compare attention scores at each stage of the BNN learning process to investigate correlations associated with BNN learning.



\bibliographystyle{unsrt}
\bibliography{main}

\appendix

\section{Background Knowledge}
\label{sec_bk}

\subsection{Biophysical Neural Network Model of the Brain}
\label{bnn}
The brain, as the most efficient yet complex information-processing system, comprises an intricate network of electrical and chemical connections between its fundamental computing units, including neurons and glial cells. These networks are responsible for processing and transmitting information throughout the body, enabling various functions such as sensation, movement, cognition, and homeostasis. However, since experimental methods \textit{in vivo} or \textit{in vitro} for investigating the brain's network structure have been limited, computational methods \textit{in silico} that simulate the brain's biological neural networks have been widely studied in the field of computational neuroscience. In this section, we will discuss the structure of biophysical neural network models of the brain.

\subsubsection{Biological Neurons and Models}
The neuron, considered the fundamental computational unit of the brain, processes incoming synaptic information received at its dendrites from other neurons in the network. When the integrated voltage exceeds a certain threshold, the neuron generates a specific pattern of spike outputs through its axon hillock. To model these neuronal dynamics, the cell membrane, a critical structure responsible for creating the relative potential between intracellular and extracellular ionic differences, can be electrophysiologically described as an electrical equivalent circuit consisting of a resistor and capacitor. Consequently, the cell membrane potential $V_m$ is represented by
\begin{equation} 
{C_m}\frac{d V_m}{d t} = - I_{ion} + I_{inj} \nonumber 
\end{equation}
where $C_m$ is the membrane capacitance, $t$ is time, $I_{ion}$ represents the total ionic currents, and $I_{inj}$ denotes the injected current. The foundational model for most biophysical computational models is the Hodgkin-Huxley (HH) theory of the action potential. Based on intracellular recordings of action potentials and ionic current measurements from squid axons, Hodgkin and Huxley formulated the action potential as resulting from a rapid inward $Na^+$ current and a more slowly activating outward $K^+$ current. The total ionic current is represented as the sum of separate $Na^+$, $K^+$, and leak currents:
\begin{subequations}
\begin{align}
 I_{ion(HH)}    & =\bar{g}_{Na}p_{Na}^{3}q_{Na}(V_m-E_{Na}) + \bar{g}_{K}p_{K}^{4}(V_m-E_{K})+\bar{g}_{Leak}(V_m-E_{Leak})\nonumber \\
\frac{d\theta}{dt}   & = \alpha_{\theta}(1-\theta)-\beta_{\theta}(\theta), ~~\theta \in \{p_{Na},q_{Na},p_{K}\} \nonumber
\end{align}
\end{subequations}
where $p_{Na}$ and $q_{Na}$ denote activation and inaction of $g_{Na}$, $p_{K}$ denotes the activation of $g_K$, $E_{{ion}\in\{Na, K, Leak\}}$ is the reversal potential of $Na^+$, $K^+$, and leak currents. 
These HH type equations can capture complex intracellular dynamics such as the shape of the action potential, threshold, refractory period, and subthreshold oscillations. Consequently, most biophysical models are based on HH type equations, with modifications for various types of ionic channels.
On the other hand, for large-scale network simulations with lower computational costs, the leaky integrate-and-fire (LIF) model can be used. This model assumes that spike generation occurs when the membrane potential exceeds a certain threshold, focusing solely on subthreshold integration.
\begin{equation}
      I_{ion(LIF)}=\bar{g}_{Leak}(V_m-E_{Leak})~~ if  ~~V_m>V_{thresh}, V_m=V_{reset} \nonumber
\end{equation}
where $V_{thresh}$ and $V_{reset}$  denotes the threshold potential and reset potential after spike generation, respectively.

\subsubsection{Multi-compartment Neurons and Models}
Although point models of spike generation, such as the HH or LIF models, can capture spiking dynamics and provide significant insights into neuronal behavior, there is a more intricate interaction between the complex spatial structures of the neuron that is crucial for neuronal information processing [ref]. Typically, neurons can be classified into three compartments: soma, dendrites, and axon, which are responsible for housing the nucleus, receiving inputs, and delivering action potentials, respectively. Therefore, modeling the spatial dynamics across these various compartments is essential for understanding how neurons process information.

In the multi-compartment model of neurons, the spatial propagation of electrical signals across these various compartments is represented by a sequence of small electrical R-C circuits that interact with each other. Along these structures, the propagation of electrical signals is computed using cable theory. Each compartment in a multi-compartment model is connected to adjacent compartments via resistances that represent the intracellular axial resistance, $R_a$, due to the cytoplasm, and each compartment is also coupled to the extracellular space through the membrane potential within compartment $i$, and $I_{inj,i}$ is the injected current into compartment $i$. The terms $\frac{V_{m,i-1} - V_{m,i}}{R_a}$ and $\frac{V_{m,i+1} - V_{m,i}}{R_a}$ represent the axial currents flowing between adjacent compartments. This allows for the simulation of complex electrophysiological dynamics at a detailed level, including dendritic integration, dendritic spikes, dendritic filtering, and synaptic plasticity, which are known to be crucial for brain information processing. 

\subsubsection{Synapse and Models}
Biological neurons transmit information through electrical or chemical synapses, with chemical synapses being the most common. In chemical synapses, the generated spike signal propagates to the axon terminal, triggering synaptic vesicle exocytosis and the subsequent release of neurotransmitters. When the released neurotransmitters cross the synaptic cleft and bind to postsynaptic receptors, postsynaptic ion channels, such as AMPA or GABA receptors, open. This alters the ionic permeability (e.g., $Na^+$, $Ca^{2+}$, or $Cl^{-}$), subsequently depolarizing or hyperpolarizing the membrane potential of the dendrites forming the synapse.

Since these steps are highly dynamic due to the chemical diffusion and reaction of neurotransmitters, a synaptic response model that evokes postsynaptic current using a unit function input without considering any synaptic current cannot accurately describe the synaptic response in the postsynaptic neuron. Moreover, even when employing a synaptic current model of a single exponential that only considers the decay phase of the postsynaptic current, it fails to fully capture the rising dynamics of the synaptic current. Therefore, in most studies, postsynaptic responses are commonly described using a double exponential, with one exponential for the rising phase and another for the decay phase of the synaptic response, as described below:
\begin{equation}\label{eq_syn_cond}
g_{syn}(t) = g_{max} (e^{-\frac{t-t_0}{\tau_{decay}}}-e^{-\frac{t-t_0}{\tau_{rise}}}) \nonumber
\end{equation}
where $g_{syn}$ is the synaptic conductance, $g_{max}$ is the maximum synaptic conductance, and $\tau_{decay}$ and $\tau_{rise}$ are the time constants of decaying and rising exponential curves, respectively. This leads synaptic current as 
\begin{equation}\label{eq_syn_cur}
I_{syn}(t) = g_{syn}(t)(V_m-E_{syn}) \nonumber
\end{equation}
where $I_{syn}$ is the synaptic current and $E_{syn}$ is the reversal potential for the synaptic current.
For the electrical synapses, it is bidirectionally connect and allow current flow between two neurons through gap junctions as typically described in
\begin{equation}
    CC = g_{j} / (g_j + g_{uninj} ) \nonumber
\end{equation}
where $CC$ is the coupling coefficient, $g_j$ is the junction conductance and  $g_{uninj}$ is the membrane conductance of the uninjected cell. 

\subsection{Self-Attention Mechanisms}
\label{sam}

We provide a brief overview of the self-attention mechanism~\citep{vaswani2017attention} and its variants. The use of self-attention is widespread, and this review is intended for those who are new to the concept.

\textbf{Self-Attention} (SA) is a method that identifies correlations between elements within feature vectors and re-represents the vectors based on those correlations.

Suppose we have a feature vector $\mathbf{f}_i = [\text{f}_{i,0}, \text{f}_{i,1}, \dots, \text{f}_{i,l-1}]^T$ of length $l$, where $0 \leq i < k$, for non-negative integers $k$ and $l$. Then a feature matrix $\mathbf{F} = [\mathbf{f}_0, \mathbf{f}_1, \dots, \mathbf{f}_{k-1}]^T \in \mathbb{R}^{k \times l}$ represents a set of feature vectors. The goal of SA is to project $\mathbf{F}$ into a more suitable vector space and find a better representation $\mathbf{F}' \in \mathbb{R}^{k \times d}$ for a positive integer $d$.

SA first computes three different matrices $\{\mathbf{Q}, \mathbf{K}, \mathbf{V}\}$ (representing query, key, and value) from $\mathbf{F}$ using $\mathbf{Q} = \mathbf{F} \cdot \mathbf{W}_Q$, $\mathbf{K} = \mathbf{F} \cdot \mathbf{W}_K$, and $\mathbf{V} = \mathbf{F} \cdot \mathbf{W}_V$. Here, the matrices $\mathbf{W}_Q$, $\mathbf{W}_K$, and $\mathbf{W}_V$ are learnable parameters, all with the same shape of $l \times d$. After obtaining $\mathbf{Q}$, $\mathbf{K}$, and $\mathbf{V}$, SA computes a score function using $\mathbf{Q}$ and $\mathbf{K}$ as arguments, producing a score matrix $\mathbf{C} \in \mathbb{R}^{k \times k}$, which represents the correlations between the vectors in $\mathbf{F}$. The score function can take various forms, with common examples including dot, scaled dot, general, concat, and location-based score functions. Finally, SA obtains $\mathbf{F}'$ by performing a softmax operation on each row of this correlation matrix and multiplying the result by $\mathbf{V}$ through matrix multiplication: $\mathbf{F}' = \text{Softmax}(\mathbf{C}) \cdot \mathbf{V}$.

In SA, the specific meanings of the latent score matrix $\mathbf{C}$ and the representation matrix $\mathbf{F}'$ depend on how $\mathbf{W}_Q$, $\mathbf{W}_K$, and $\mathbf{W}_V$ are learned and on the score function used.

\textbf{Masked Self-Attention} (MSA) is used when certain vectors within $\mathbf{F}$ do not have a clear correlation in any direction, or when the user of SA intentionally wants to exclude correlations between specific vectors. MSA adds an additional step where a value of $-\infty$ is applied to the score matrix $\mathbf{C}$, obtained from the SA score function, to mask the correlations that should be excluded.

\textbf{Multi-Head Self-Attention} (MHSA) refers to performing MSA multiple times in parallel. Specifically, for a positive integer $n_{head}$, this means using $n_{head}$ different sets of $\{\mathbf{W}_Q, \mathbf{W}_K, \mathbf{W}_V\}$ to generate $n_{head}$ different sets of $\{\mathbf{Q}, \mathbf{K}, \mathbf{V}\}$. The result of MHSA is obtained by concatenating the $n_{head}$ resulting matrices from each MSA.

\textbf{Grouped Multi-Query Self-Attention} (GMQSA)~\citep{ainslie2023gqa} is a variation of MHSA. While MHSA generates and uses $n_{head}$ different sets of $\{\mathbf{Q}, \mathbf{K}, \mathbf{V}\}$, GMQSA generates $n_{query}$ sets of $\{\mathbf{K}, \mathbf{V}\}$ for a positive integer $n_{query}$ less than $n_{head}$, and shares these $n_{query}$ sets among $n_{head}$ different $\mathbf{Q}$ in a grouped manner.

\textbf{Multi-Query Self-Attention} (MQSA)~\citep{shazeer2019fast} is another variation of MHSA. While MHSA generates and uses $n_{head}$ different sets of $\{\mathbf{Q}, \mathbf{K}, \mathbf{V}\}$, MQSA generates a single set of $\{\mathbf{K}, \mathbf{V}\}$, which is shared among $n_{head}$ different $\mathbf{Q}$.

\textbf{Hierarchical Attention} (HA)~\citep{yang2016hierarchical} gathers data at different levels of abstraction within the input sequence, organizing information hierarchically and aggregating data at various levels, such as word-level and sentence-level representations.

\subsection{Computational Graphs}

A computational graph (CG), denoted as $G(U,E)$, is a graphical representation of a computational process. It consists of nodes $U_i \in U$, where each node represents either a mathematical operation or a variable, and directed edges $E_{ij} \in E$, which represent the dependencies and flow of data or information between two nodes, $U_i$ and $U_j$, for non-negative integers $i$ and $j$.

In a CG, operation nodes perform mathematical functions such as addition, multiplication, transformations, or kernel functions. 
Variable nodes, on the other hand, store parameters, inputs, or intermediate values passed through the graph.

Each CG node has associated features $\mathbf{f}_i\in\mathbb{R}^l$, which capture attributes or data related to the node $U_i$, where $l$ is the number of features.
These features may include categorical, numerical, vectorial, or binary information.
For example, $\mathbf{f}_i$ can represent the type of operation or variable, the format of input/output data, execution time, memory usage, schedulability, or the distance (in terms of the number of hops) from other nodes.

Processing CGs fundamentally relies on graph processing methodologies, as CGs belong to a category of graphs. Examples of such methodologies include Graph Convolutional Networks~\citep{kipf2017semi}, GraphSAGE~\citep{hamilton2017inductive}, Transformers~\citep{vaswani2017attention}, Graph Attention Networks~\citep{velivckovic2018graph}, and GFlowNets~\citep{zhang2023robust}.

\textbf{GraphSAGE} is a graph neural network (GNN) model specifically designed for learning node representations in large graphs. In \citep{zhou2020transferable}, it is used to analyze and project the CG of a machine learning model into a lower dimension. By capturing both local and global structures, GraphSAGE operates inductively, allowing it to generalize to nodes that were not seen during training. GraphSAGE functions through two key stages: sampling and aggregation. In the sampling stage, a fixed-size set of neighboring nodes is selected for each target node within the graph. Subsequently, in the aggregation stage, GraphSAGE aggregates and summarizes information from the sampled neighbors to learn the node’s representation.

\textbf{Transformer} introduces (masked) multi-head self-attentions (MHSA) to manage dependencies between tokens (or nodes). MHSA is adopted in \citep{gagrani2022neural} to achieve a topological ordering for CGs. This methodology receives structural information from seven different versions of the CG transformation, with each version encompassing various topological features of the operand nodes and edges. To understand the relationships and information flow between nodes within each CG, each layer of the encoder utilizes seven MHSAs. Additionally, a feed-forward neural network is employed to update and capture the features of each operand node.

\textbf{Graph Attention Network} (GAT) is a class of neural networks specifically designed to process graph-structured data. Unlike GNNs, which utilize fixed aggregation schemes such as mean or sum to integrate information from neighboring nodes, GAT incorporates an attention mechanism that dynamically assigns varying levels of importance to each neighboring node during the message-passing process. This adaptive weighting allows GAT to focus on the most relevant nodes and edges, thereby enhancing its capacity to capture complex, correlations in graph-structured data.

In GAT, attention coefficients are computed for each edge by comparing the features of the connected nodes. 
The process begins by applying a linear transformation to each node’s feature vector $\mathbf{f}_i$ using a shared weight matrix $\mathbf{W}\in\mathbb{R}^{k \times l}$, yielding a transformed feature vector $\mathbf{f}_i'=\mathbf{W}\cdot\mathbf{f}_i$ where $\mathbf{f}_i'\in \mathbb{R}^k$, $k$ is the dimension of the transformed features, and $\cdot$ denotes matrix multiplication.
Subsequently, attention coefficients are computed to quantify the importance of each neighboring node’s features.
For a pair of connected nodes $U_i$ and $U_j$, the unnormalized attention coefficient $e_{ij}$ is defined as $e_{ij}=\sigma(\mathbf{a}^T[\mathbf{f}_i' || \mathbf{f}_j'])$ where $\sigma$ is an activation function, $\mathbf{a} \in \mathbb{R}^{2k}$ is a learnable weight vector, and $||$ denotes concatenation.
The unnormalized attention coefficients $e_{ij}$ are then normalized via the softmax function, resulting in normalized attention scores, which sum to 1 across all neighbors of node $U_i$.

These normalized attention scores are subsequently employed as weights to aggregate information from neighboring nodes, determining the contribution of each neighboring node’s features to the updated $\mathbf{f}_i$.
This selective amplification or attenuation of neighboring nodes’ influence empowers GAT to effectively model varying interactions and dependencies across the graph. 
Moreover, the attention mechanism can be applied either at the node level, aggregating information from neighboring nodes, or at the edge level, directly modeling the relationships between connected nodes.

\end{document}